\documentclass[letterpaper, 10 pt, conference]{ieeeconf}  

\IEEEoverridecommandlockouts                              

\usepackage{cite}
\usepackage{xcolor}
\usepackage{amsmath}
\usepackage{siunitx}
\usepackage{textcomp}
\usepackage{float}
\usepackage{mwe}
\usepackage{booktabs}
\usepackage{graphicx}
\usepackage{subfig}
\usepackage{longtable}
\usepackage{amsfonts}
\usepackage{makecell}
\usepackage{changepage}
\usepackage{multicol}
\usepackage{lipsum}
\usepackage{mwe}
\usepackage{hyperref}
\usepackage{balance}

\title{\LARGE \bf
Deep Camera Pose Regression Using Pseudo-LiDAR
}

\author{Ali Raza$^{1}$, Lazar Lolic$^{1}$, Shahmir Akhter$^{1}$, Alfonso Dela Cruz$^{1}$, and Michael Liut$^{1}$  
\thanks{$^{1}$Department of Mathematical and Computational Sciences, University of Toronto Mississauga, Ontario, Canada. Emails: {\tt\small \{alisyed.raza, lazar.lolic, shahmir.akhter, alfonso.delacruz\}@mail.utoronto.ca}, and {\tt\small michael.liut@utoronto.ca}.
}}

\begin{document}

\thispagestyle{plain}
\pagestyle{plain}
\maketitle


\begin{abstract}
An accurate and robust large-scale localization system is an integral component for active areas of research such as autonomous vehicles and augmented reality. To this end, many learning algorithms have been proposed that predict 6DOF camera pose from RGB or RGB-D images. However, previous methods that incorporate depth typically treat the data the same way as RGB images, often adding depth maps as additional channels to RGB images and passing them through convolutional neural networks (CNNs). In this paper, we show that converting depth maps into pseudo-LiDAR signals, previously shown to be useful for 3D object detection, is a better representation for camera localization tasks by projecting point clouds that can accurately determine 6DOF camera pose. This is demonstrated by first comparing localization accuracies of a network operating exclusively on pseudo-LiDAR representations, with networks operating exclusively on depth maps. We then propose FusionLoc, a novel architecture that uses pseudo-LiDAR to regress a 6DOF camera pose. FusionLoc is a dual stream neural network, which aims to remedy common issues with typical 2D CNNs operating on RGB-D images. The results from this architecture are compared against various other state-of-the-art deep pose regression implementations using the 7 Scenes dataset. The findings are that FusionLoc performs better than a number of other camera localization methods, with a notable improvement being, on average, 0.33m and \textbf{4.35$^{\circ}$} more accurate than RGB-D PoseNet. By proving the validity of using pseudo-LiDAR signals over depth maps for localization, there are new considerations when implementing large-scale localization systems.\\
\end{abstract}

\section{Introduction}

The challenge of creating a large scale localization system has been a thriving area of research for many years. Obtaining accurate pose information is crucial for many applications such as autonomous vehicles, mobile robotics, and augmented reality. Although the sensors used for localization vary depending on the application and devices involved, cameras are by far the most common given their cost, ease of integration, and quality of output. Cameras prove to be useful for a range of localization tasks, from large-scale localization in autonomous vehicles \cite{self-driving_survey} to smartphone-based localization in indoor environments \cite{2021razaetal}.

Many camera localization systems have been proposed in the past. This includes systems that use feature matching like SIFT \cite{SIFT}, SURF \cite{SURF}, or ORB \cite{ORB} which prove to be effective for visual odometry. However, performance deteriorates when deployed for large-scale global pose estimation. The features are inconsistent across varying lighting and weather conditions and fail to be accurately captured for repeating patterns or texture-less areas, typically found in indoor environments. 

In recent years, deep learning-based camera pose regression techniques have shown promise. Kendall \textit{et al.} \cite{kendall2016posenet} showed that a 6DOF Pose can be regressed from a single RGB image using a convolutional neural network (CNN). Such networks learn useful pose information in their feature vectors without requiring any hand-crafting. Li \textit{et al.} \cite{dualstream_posenet} followed up by showing that depth information can be incorporated to improve the performance of such convolutional neural networks. However, depth maps in such approaches are treated the same way as RGB images. In particular, they are typically stacked as multiple channels and passed through CNNs that are pretrained on image datasets like ImageNet \cite{ImageNet}. Since depth maps are dense representations containing per-pixel depth, convolving them loses a lot of captured information that may be valuable for localization (discussed further in \hyperref[sec:experiment]{Section IV.C}). Recently Wang \textit{et al.} \cite{wang2020pseudolidar} proposed a system called pseudo-LiDAR to extrapolate point clouds from depth images. These point clouds can then be used for 3D object detection and segmentation using algorithms like PointNet \cite{qi2017pointnet}.
While pseudo-LiDAR was designed to be used for 3D object detection, we believe it is a representation that can be used to accurately determine a 6DOF camera pose from a single RGB-D image. Thus, our goals in this paper are as follows:
\begin{enumerate}
    \item to prove the viability of pseudo-LiDAR within this space; and
    \item to propose a novel architecture that uses pseudo-LiDAR to regress a 6DOF camera pose. 
\end{enumerate}

\begin{figure*}[hbt!]
 \centering
  \includegraphics[width=.9\linewidth]{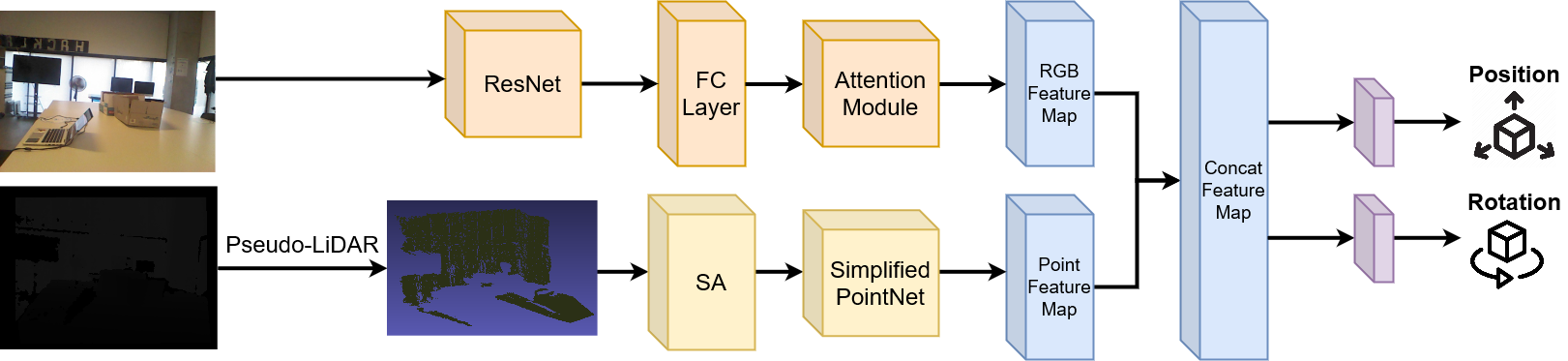}
  \renewcommand{\thefigure}{1}
  \captionsetup{font=small}
  \captionof{figure}{FusionLoc architecture diagram}
  \label{fig:diagram}
\end{figure*}

We address the first point by presenting a simple pose regression architecture inspired by PointNet. This architecture operates on point sets generated using pseudo-LiDAR and compares it against a typical pose regression system operating exclusively on depth maps. Drawing inspiration from the work of Wang \textit{et al.}, we also visually show the information lost when convolving depth maps using images from a popular localization dataset, 7 Scenes \cite{7scenes}. \\
\indent Ultimately, we propose FusionLoc, a dual stream neural network which addresses common issues with localization CNNs operating on RGB-D images. In FusionLoc, one stream processes RGB images while the second uses point sets generated by converting depth maps using pseudo-LiDAR as input. The features learned from both are finally concatenated and passed through fully connected layers that predict the pose. FusionLoc is agnostic to different depth estimation algorithms. Therefore, it can operate on depth maps generated using stereo depth estimation techniques \cite{stereo1, stereo2}, monocular depth estimation \cite{mono1, mono2}, and infrared cameras such as the ones found in Kinect sensors.



\section{Related Work}

Image-based localization in large scale environments can be categorized into two groups; place recognition and metric localization. 
\subsection{Place Recognition}
Place recognition typically uses image retrieval techniques. This involves matching queried images with images from a database containing known pose information, using approaches such as Bag-of-Words (BoW) \cite{bow} or Vector of Locally Aggregated Descriptors (VLAD) \cite{VLAD} to estimate a camera pose for the queried image. Descriptors can also be learned through deep learning. However, while these methods scale very well to large environments, the pose estimate is coarse and not fine grained. Image retrieval has also been used as part of structure-based approaches that involve 2D to 3D correspondence \cite{image-retrieval2}.

\subsection{Metric Localization}

\subsubsection{Feature Matching} Another solution involves performing 2D-3D correspondence between a queried image and a 3D point cloud of the environment to recover the camera pose. Using Structure-from-Motion (SfM), a sparse 3D point cloud of the environment is created where each point has associated feature descriptors. These descriptors can be matched with sparse features of the queried image to establish 2D-3D correspondence that produce very accurate pose estimates when used with RANSAC-based pose optimization techniques~\cite{2d_3d_1, 2d_3d_2, 2d_3d_3}. Techniques have also been introduced to make descriptor matching more efficient for real-time pose estimation \cite{efficient_2d_3d} and scale better to large city-scale scenes \cite{worldwide_2d_3d, worldwide_2d_3d_2}. However, feature detectors often fail to find points when motion blur or texture-less surfaces are involved. SfM reconstructions are also not very accurate for scenes where there may be many texture-less surfaces or repeating patterns, commonly found in indoor spaces.

\subsubsection{Deep Pose Regression}
PoseNet, the pioneering work of Kendall \textit{et al.}, showed that deep learning can be used for metric localization through a convolutional neural network that could directly regress camera pose from input RGB images \cite{kendall2016posenet}. PoseNet uses networks like ResNet \cite{resnet} and GoogleNet \cite{googlenet}, pre-trained on large-scale image classification data as feature extractors, which are then followed by linear regression layers that output 6DOF pose. Unlike image retrieval and feature matching techniques, this approach does not require a feature dataset or a 3D model that requires memory linearly proportional to the size of the environment. However, the accuracy is not comparable to that produced by the aforementioned methods. \\
\indent Deep learning-based camera relocalization was further improved in followup work by Kendall \textit{et al.} with Bayesian neural networks \cite{bayesian_posenet} and with loss functions that leverage geometry \cite{geometric_posenet}. Walch \textit{et al.} incorporated LSTMs \cite{lstm_posenet} to reduce feature dimensionality in a structured way, and Clark \textit{et al.} proposed a recurrent model for pose estimation over videos \cite{clark2017vidloc}. Li \textit{et al.} introduced a dual stream version of PoseNet to use RGB-D data \cite{dualstream_posenet}. Wang \textit{et al.} propose using attention to force the network to focus on more geometrically robust changes \cite{wang2019atloc}.

\subsection{Deep Learning for Point Clouds}
Deep Learning has shown promise for tasks using point clouds such as 3D object detection, part segmentation, and semantic segmentation. Qi \textit{et al.} proposed PointNet, a deep neural network that learns features directly from point sets~\cite{qi2017pointnet}. Follow-up work further improved learning feature embeddings such as PointNet++, which learned local structures in point sets, showing improvements on object detection and segmentation tasks \cite{qi2017pointnetplusplus}. Zhou \textit{et al.} also proposed VoxelNet, an end-to-end network that divides point clouds into voxels for object detection \cite{zhou2017voxelnet}. \\




\section{Implementation}

\subsection{Pseudo-LiDAR}

Wang \textit{et al.} proposed an approach to lessen the gap between image and LiDAR-based 3D object detection \cite{wang2020pseudolidar}. They argued that a large reason for the discrepancy was the representation. In LiDAR based point clouds, object shapes and sizes are maintained regardless of the depth. When such point clouds are processed, these properties are preserved and all objects are treated the same way regardless of the depth. However, when using RGB-D data, most techniques rely on 2D convolution and typically stack multiple RGB and depth channels to leverage transfer learning, using pre-trained image classification networks. \\
\indent The issue arises when we consider 2D convolutions which consider local patches to contain valuable information and that all pixel neighbourhoods can be operated on in a similar manner. This presents issues when dealing with depth maps. For example, when considering patches with more than one object, pixels for each object may be far away from each other. Moreover, objects with varying depth will be scaled accordingly in the depth map. Objects far away will appear much smaller compared to objects close by which may be treated differently. \\
\indent While pseudo-LiDAR was proposed for 3D object detection, we argue that a similar argument can be made for camera relocalization. A large part of predicting a 6DOF pose involves knowing how far the camera is in reference to the objects observed in the image. When different objects appear together across varying images, pixels corresponding to each object may appear together in local patches. However, depending on where the camera is located, one object may be closer to the camera than the others, thus producing a completely different pose.\\ 

\subsubsection{Pseudo-LiDAR Implementation} Pseudo-LiDAR allows us to obtain 3D points ${x,y, z}$ for each pixel $(u, v)$ in depth maps using the following:
\begin{equation}
    z = D(u,v)
\end{equation}
\begin{equation}
    x = \frac{z (u - c_u)}{f_u} 
\end{equation}
\begin{equation}
    y = \frac{z (v - c_v)}{f_v} 
\end{equation}
where $D(u,v)$ represents the depth of the pixel, $(u, v)$, $(c_u, c_v)$ represent the camera's centre pixel, and $f_u$ and $f_v$ represent the horizontal and vertical focal lengths respectively \cite{wang2020pseudolidar}. 

\subsection{FusionLoc Architecture}
\label{sec:architecture}
Our network architecture, FusionLoc (shown in \hyperref[fig:diagram]{Fig. 1}), is a dual stream neural network, with one stream processing RGB images, while the second uses point sets generated using pseudo-LiDAR as input. The features learned from both are then concatenated and passed through fully connected layers that predict the pose.\\

\subsubsection{RGB Stream}
The RGB stream follows a convention similar to other camera pose regression papers like PoseNet \cite{kendall2016posenet} and Hourglass-Pose \cite{hourglass}. It contains a visual encoder which is followed by a self-attention layer, given its benefits showcased in recent works \cite{wang2019atloc}.
\paragraph{Visual Encoder}
The visual encoder serves as a feature extractor that takes RGB images as input and extracts features that are useful for localization. Due to limited training data, it is better to use transfer learning instead of training convolutional neural networks from scratch \cite{kendall2016posenet}. Kendall \textit{et al.} showed the value of using pre-trained networks for camera localization \cite{kendall2016posenet}, which has since then been adopted by most camera pose regression architectures. We take a similar approach by using ResNet34 \cite{resnet} as our visual encoder which has been shown to be more stable than other architectures \cite{brahmbhatt2018geometryaware}. The final softmax layer is replaced with a fully connected layer of size 1024 to further learn localization features. 

\paragraph{Attention Module}
Previous works have shown self-attention modules to be useful in focusing on geometrically useful features and ignoring noise. We adopt a self-attention module proposed by Wang \textit{et al.} \cite{wang2019atloc} which attempts to generate feature maps that capture global correlations from separated spatial regions in images.
The features $x\in\mathbb{R}^{C}$ obtained using Resnet34 are first used to obtain embeddings $\theta(x_i) = W_{\theta}x_i$ and $\phi(x_j) = W_{\phi}x_j$. These embeddings are then dotted and normalized to obtain attention vectors that can be written as:
\begin{equation}
    y = Softmax(x^T\mathbb{W}_{\theta}^T\mathbb{W}_\phi)\mathbb{W}_gx
\end{equation}
where $\mathbb{W}_gx$ is another linear transform. Wang \textit{et al.}'s work contains a detailed description on this attention module \cite{wang2019atloc}.\\

\subsubsection{Point cloud Stream}
\paragraph{Set Abstraction}
We adopt set abstraction layers from PointNet++ \cite{qi2017pointnetplusplus} which are made of three key layers: Sampling layers, Grouping layers, and PointNet layers. Sampling layers use iterative farthest point to sample subset of points, which typically has better coverage of the entire point set compared to random sampling.
Grouping layers take point sets of size $N \times (d + C)$ and coordinates of centroids of size $N' \times d$, and output a group of point sets $N'\times K \times (d + C)$ where each group is a local region and $K$ is the number of points in the neighbourhood of centroid points. The final PointNet layer outputs feature vectors that encode point-to-point relations in local regions \cite{qi2017pointnetplusplus}.
\paragraph{Simplified PointNet}
We follow the set abstraction layers with a structure inspired by PointNet \cite{qi2017pointnet}. PointNet can be described as a universal approximator that can be represented using:
\begin{equation}
    f(x_1, ..., x_n) = \theta(\textbf{MAX}[h(x_i) | x_i \in \textbf{P}])
\end{equation}
where \textbf{MAX} is maximum pooling, $h(x)$ and $\theta(x)$ are multi-layer perceptrons (MLPs), and \textbf{P} is the input point cloud.
However, unlike PointNet, our model does not contain input or feature transform modules (T-Net). The purpose of T-Net is to make models transform invariant so that rotations and translations do not change the point cloud category. While this is important for object detection and segmentation tasks, it is trivial that such layers significantly undermine localization performance which would be required to output a pose, taking the position and orientation of the point cloud in consideration.
\subsubsection{Final layers}
The final layer concatenates feature maps obtained from both streams and passes the concatenated feature map through fully connected layers that output the position and orientation.
\subsection{Loss Function}
Given an image \textbf{I}, we aim to predict a 6DOF camera pose [\textbf{t}, \textbf{q}], where \textbf{t} is the position and \textbf{q} is a unit quaternion for rotation. We optimize using the following loss function:
\begin{equation}
    loss(I) = {\lVert \bf{p} - \bf{\overline{p}} \rVert}_1  e^{- \beta} + \beta + \lVert \log\bf{q}  - \log \bf{\overline{q}} \rVert e^{- \gamma} + \gamma
\end{equation}
as shown by Brahmbhatt \textit{et al.} \cite{brahmbhatt2018geometryaware},
where ${\beta}$ and ${\gamma}$ are learnable parameters to learn translation and rotation. $\log\textbf{q}$ is the logarithmic form of a unit quaternion defined as the following:

\begin{equation}
    \bf{q} = 
    \begin{cases} 
      \frac{\bf{v}}{\lVert \bf{v} \rVert}\cos^{-1}u & \text{if } \lVert \bf{v} \rVert \neq 0\\
      0 & \text{otherwise}
   \end{cases}
\end{equation}
where the quaternion is $\bf{q}=(u,\bf{v})$, u is the scalar (real) part of the quaternion and v is the 3D vector (imaginary) part.\\

\section{Evaluation}

In order to showcase the effectiveness of pseudo-LiDAR, we first compare a depth only PoseNet implementation, with a network comprising of a PointNet architecture used for pose regression that is given point sets generated using pseudo-LiDAR as input. We then evaluate our proposed architecture in \hyperref[sec:architecture]{Section III.B} by comparing it with state-of-the-art deep pose regression approaches.

\subsection{Dataset}
We used the 7 Scenes dataset \cite{7scenes} for all our experiments, which contains RGB-D images for seven different indoor locations captured using a handheld first generation Microsoft Kinect with the ground truth camera poses calculated using KinectFusion. The images have a resolution of 640×480 pixels with training and testing sequences provided. The dataset is popular for visual relocalization, as it contains scenes with varying conditions and camera effects such as motion blur and texture-less surfaces, which are challenging for the given task.

\subsection{Setup}
For images, we downscale the 640x480 images so that the smallest dimension is 256. These images are then cropped into a 224x224 image and normalized. For training, they are randomly cropped, and for testing, they are cropped from the centre. For our RGB stream, since we use a ResNet34\cite{resnet} model that was pre-trained on the ImageNet dataset \cite{ImageNet}, we normalize using the mean and standard deviation specified for ImageNet. For the point set stream, we project point clouds from the depth maps using pseudo-LiDAR and randomly sample 1024 points.\\
\indent For training our networks, we used the Adam optimizer \cite{adam} with a learning rate of $1 \times 10^{-4}$ and weight decay of $5 \times 10^{-4}$. The networks were trained to convergence using a batch size of 64 on an RTX 3090.

\subsection{Pseudo-LiDAR vs Depth Maps for Localization}
\label{sec:experiment}
In order to show the superior performance of using pseudo-LiDAR compared to using depth maps for localization, we compare the results of the two approaches. As previously mentioned, in LiDAR based point clouds, object shapes and sizes are maintained regardless of depth. This will preserve more information, typically leading to more accurate results.\\
\indent In order to show this, we present and compare two methods operating exclusively on depth information, in \hyperref[tab:table2]{Table I}. The first is depth-only PoseNet, which follows the depth stream from RGB-D PoseNet\cite{dualstream_posenet}, where the depth map has a jet colormap applied to create RGB channels that are fed into a ResNet based visual encoder. This is followed by a fully connected layer to finally output the pose. \\
\indent The second method, PointNet-Pose, converts depth maps into pseudo-LiDAR signals and passes them through a network that is the same as the simplified PointNet architecture used in FusionLoc, followed by a fully connected layer to output the pose.

\begin{table}[H]
\label{tab:table2}
\begin{center}
\begin{tabular} {l|l|l}
\hline
Scene               & Depth-only PoseNet                & PointNet-Pose \\ \hline
Chess               & 0.33m, \ang{13.1}                 & \textbf{0.19m}, \textbf{5.22$^{\circ}$}\\
Fire                & 0.53m, \ang{19.9}                 & \textbf{0.46m}, \textbf{14.2$^{\circ}$}\\
Heads               & \textbf{0.23m}, \ang{14.5}        & 0.35m, \textbf{11.2$^{\circ}$}\\
Office              & 0.43m, \ang{18.0}                 & \textbf{0.36m}, \textbf{13.4$^{\circ}$}\\
Pumpkin             & 0.44m, \ang{10.0}                 & \textbf{0.34m}, \textbf{8.36$^{\circ}$}\\
Red Kitchen         & 0.69m, \ang{17.0}                 & \textbf{0.32m}, \textbf{10.5$^{\circ}$}\\
Stairs              & 0.51m, \textbf{14.4$^{\circ}$}    & \textbf{0.44m}, \ang{16.1}\\ \hline\hline
Average             & 0.45m, \ang{15.3}                 & \textbf{0.35m}, \textbf{11.3$^{\circ}$}\\
\end{tabular}
\captionsetup{font=small}
\caption{Comparing depth-only PoseNet with PointNet-Pose.}
\end{center}
\end{table}

\begin{figure}[H]
    \begin{center}
    \hspace{-3pt}\includegraphics[scale=0.15]{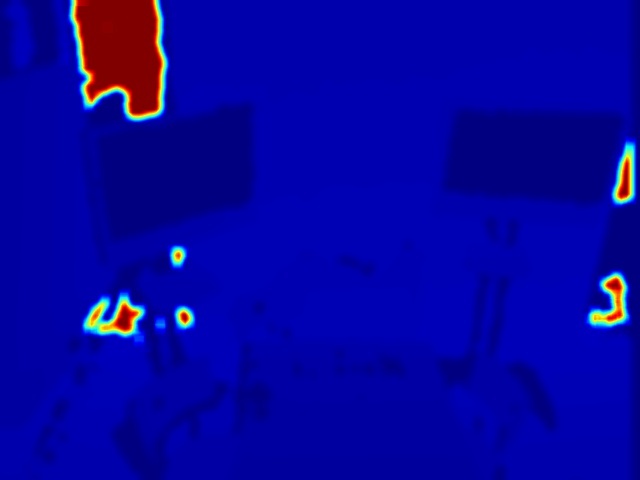}
    \includegraphics[scale=0.15]{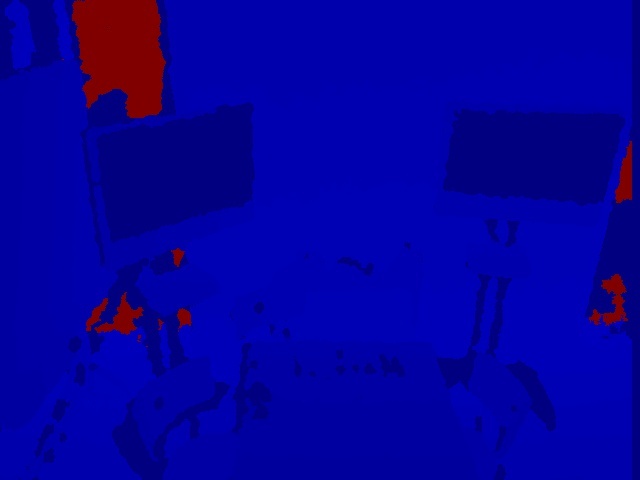}\\ \vspace{3pt}
    \includegraphics[scale=0.6]{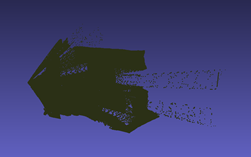}
    \includegraphics[scale=0.6]{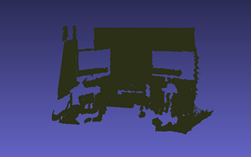}
    \renewcommand{\thefigure}{2}
    \captionsetup{font=small}
    \caption{Convolved depth map (top-left). Raw depth map (top-right). Point cloud from convolved map (bottom-left). Point cloud from raw depth map (bottom-right).}
    \label{fig:experiment}
    \end{center}
\end{figure}

Furthermore, in order to visualize the effect of applying 2D convolutions on depth maps, in \hyperref[fig:experiment]{Fig. 2} we perform an experiment similar to one done by Wang \textit{et al.} \cite{wang2020pseudolidar}. We take the original depth map and apply a single 11x11 2D convolution on it. We then convert the new depth map to a pseudo-LiDAR representation and visually compare it with the pseudo-LiDAR representation of the original depth map.


\begin{table*}[ht!]
\begin{adjustwidth}{-1.5cm}{-1cm} 
\label{tab:results}
\begin{center}
\begin{tabular}{l|llllll|lll}

\hline

Scene & 
\thead{PoseNet \\ ICCV' 15} & 
\thead{Bayesian \\ PoseNet} & 
LSTM-Pose & 
VidLoc & 
\thead{Hourglass- \\ Pose} & 
\thead{RGB-D \\ PoseNet}  &
\thead{Depth-only \\ PoseNet} &
PointNet-Pose &
FusionLoc \\ 

\hline

Chess   & 0.32m, \ang{8.12}     & 0.37m, \ang{7.24}     & 0.24m, \ang{5.77} 
        & 0.18m, N/A            & 0.15m, \ang{6.53}     & 0.36m, \ang{6.91} 
        & 0.33m, \ang{13.1}     & 0.19m, \ang{5.22}      & 0.14m, \ang{4.79}\\

Fire    & 0.47m, \ang{14.4}     & 0.43m, \ang{13.7}     & 0.34m, \ang{11.9} 
        & 0.26m, N/A            & 0.29m, \ang{11.6}    & 0.56m, \ang{13.6} 
        & 0.53m, \ang{19.9}     & 0.46m, \ang{14.2}     & 0.31m, \ang{10.7}\\
        
Heads   & 0.29m, \ang{12.0}     & 0.31m, \ang{12.0}     & 0.21m, \ang{13.7}             & 0.14m, N/A            & 0.21m, \ang{14.5}    & 0.39m, \ang{15.3} 
        & 0.23m, \ang{14.5}     & 0.35m, \ang{11.2}    & 0.15m, \ang{13.0}\\

Office  & 0.48m, \ang{7.68}     & 0.48m, \ang{8.04}     & 0.30m, \ang{8.08}             & 0.26m, N/A            & 0.21m, \ang{9.25}     & 0.63m, \ang{12.6} 
        & 0.43m, \ang{18.0}     & 0.36m, \ang{13.4}     & 0.22m, \ang{6.13}\\

Pumpkin & 0.47m, \ang{8.42}     & 0.61m, \ang{7.08}     & 0.33m, \ang{7.00}             & 0.36m, N/A            & 0.27m, \ang{6.93}     & 0.51m, \ang{8.43} 
        & 0.44m, \ang{10.0}     & 0.34m, \ang{8.36}     & 0.21m, \ang{3.83}\\

Red Kitchen & 0.59m, \ang{8.64} & 0.58m, \ang{7.54}     & 0.37m, \ang{8.83}                 & 0.31m, N/A        & 0.27m, \ang{9.82}     & 0.94m, \ang{18.2} 
            & 0.69m, \ang{17.0} & 0.32m, \ang{10.5}     & 0.24m, \ang{7.14}\\

Stairs      & 0.47m, \ang{13.8} & 0.48m, \ang{13.1}     & 0.40m, \ang{13.7}                 & 0.26m, N/A        & 0.29m, \ang{13.1}     & 0.53m, \ang{11.9} 
            & 0.51m, \ang{14.4} & 0.44m, \ang{16.1}     & 0.35m, \ang{11.0}\\

\hline\hline

Average     & 0.44m, \ang{10.4} & 0.47m, \ang{9.81}     & 0.31m, \ang{9.85}                 & 0.25m, N/A        & 0.24m, \ang{10.24}    & 0.56m, \ang{12.41} 
            & 0.45m, \ang{15.3} & 0.35m, \ang{11.3} & 0.23m, \ang{8.06}

\end{tabular}
\end{center}
\end{adjustwidth}
\captionsetup{font=small}
\caption{Comparing results from various techniques on the 7 Scenes dataset.}
\end{table*}

\begin{figure*}[!hbt]
        \includegraphics[width=.14\textwidth]{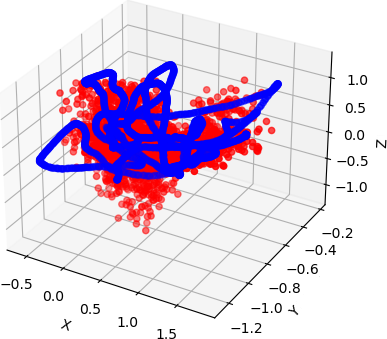}\hfill
        \includegraphics[width=.14\textwidth]{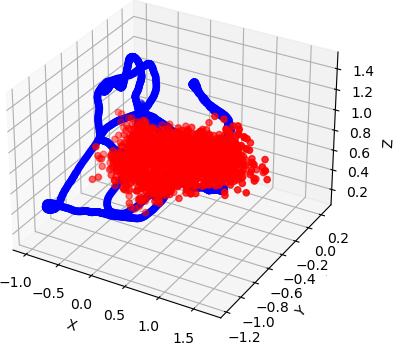}\hfill
        \includegraphics[width=.14\textwidth]{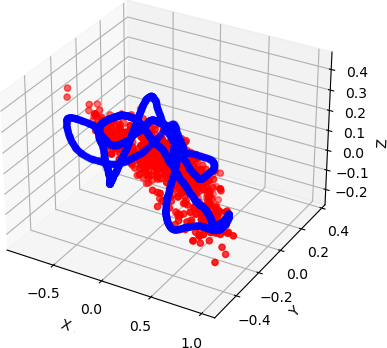}\hfill
        \includegraphics[width=.14\textwidth]{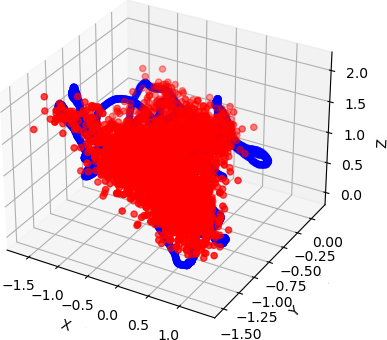}\hfill
        \includegraphics[width=.14\textwidth]{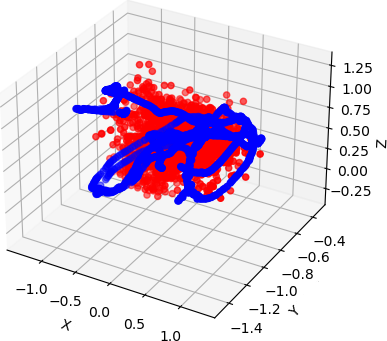}\hfill
        \includegraphics[width=.14\textwidth]{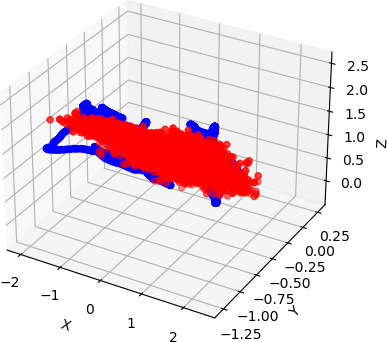}\hfill
        \includegraphics[width=.14\textwidth]{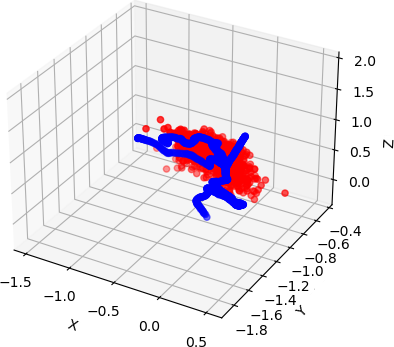}
        \captionsetup{font=small}
        \caption{Depth-only PoseNet results}
        \label{fig:graphs}
\end{figure*}
\begin{figure*}[!hbt]
        \includegraphics[width=.14\textwidth]{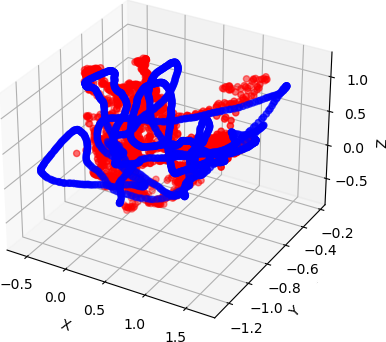}\hfill
        \includegraphics[width=.14\textwidth]{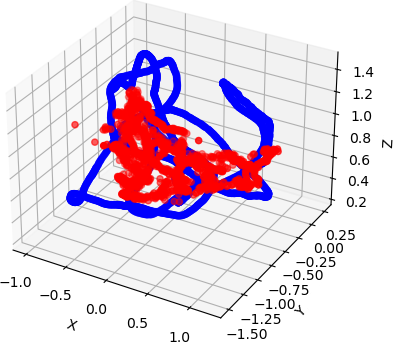}\hfill
        \includegraphics[width=.14\textwidth]{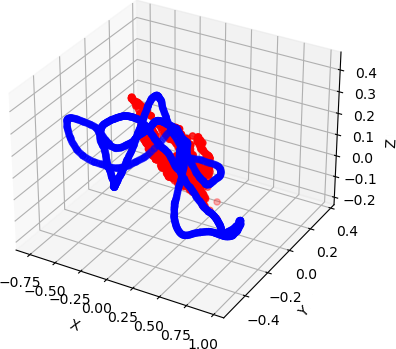}\hfill
        \includegraphics[width=.14\textwidth]{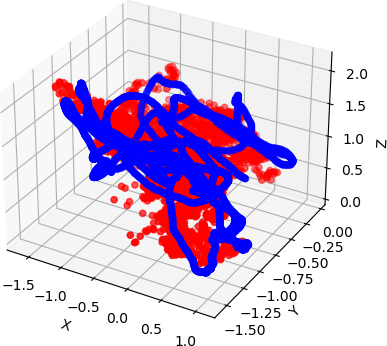}\hfill
        \includegraphics[width=.14\textwidth]{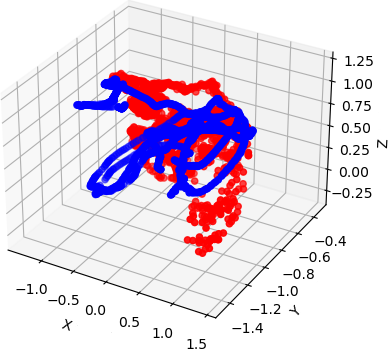}\hfill
        \includegraphics[width=.14\textwidth]{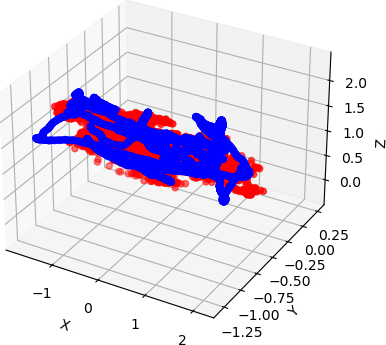}\hfill
        \includegraphics[width=.14\textwidth]{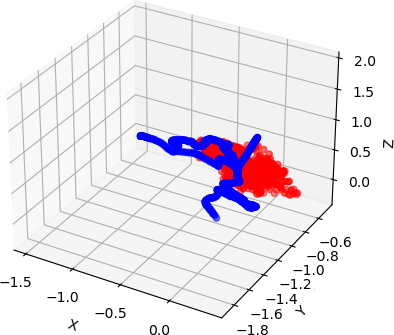}
        \captionsetup{font=small}
        \caption{PointNet-Pose results}
\end{figure*}
\begin{figure*}[!hbt]
        \includegraphics[width=.14\textwidth]{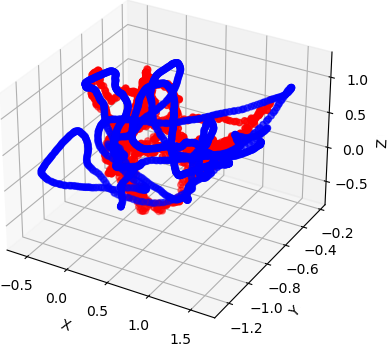}\hfill
        \includegraphics[width=.14\textwidth]{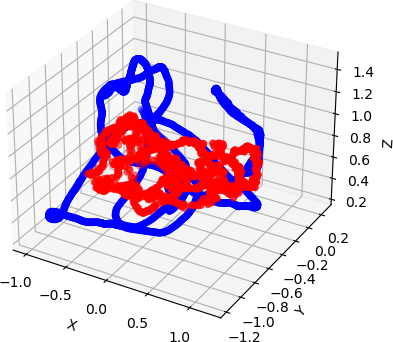}\hfill
        \includegraphics[width=.14\textwidth]{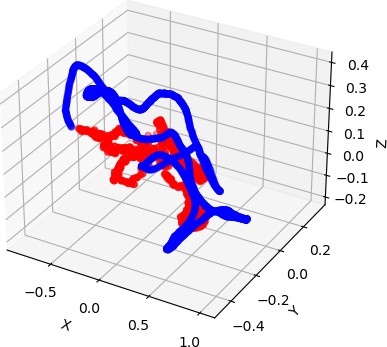}\hfill
        \includegraphics[width=.14\textwidth]{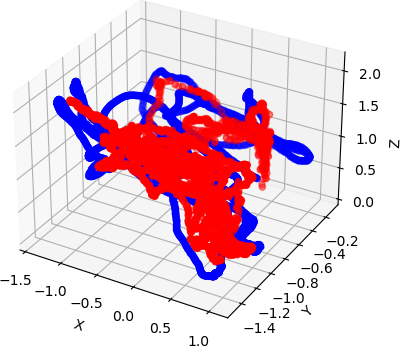}\hfill
        \includegraphics[width=.14\textwidth]{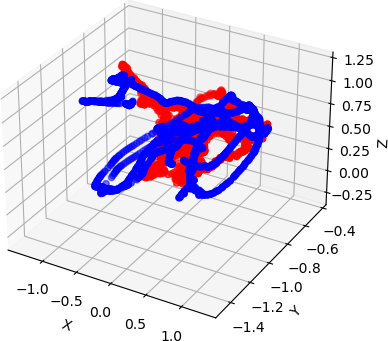}\hfill
        \includegraphics[width=.14\textwidth]{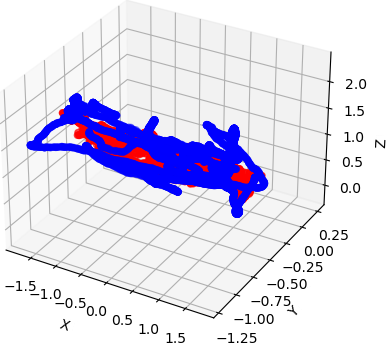}\hfill
        \includegraphics[width=.14\textwidth]{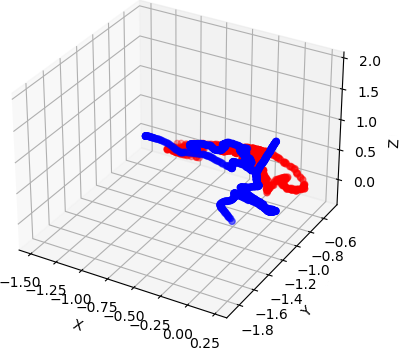}
    \captionsetup{font=small}
    \caption{FusionLoc results}
\end{figure*}

As we can see in \hyperref[fig:experiment]{Fig. 2}, the points in the point cloud generated from the convolved depth map are stretched out beyond their proper locations, while the points in the point cloud generated from the original depth map replicate a LiDAR signal for the image more accurately. This shows that important properties are preserved with the pseudo-LiDAR point cloud. This is what we feed into our network.

\subsection{Baseline}
To validate the performance of our network, we compare the results to those of various competing and state-of-the-art approaches. The methods we choose are PoseNet \cite{kendall2016posenet}, Bayesian PoseNet \cite{bayesian_posenet}, LSTM-Pose \cite{lstm_posenet}, VidLoc \cite{clark2017vidloc}, Hourglass-Pose \cite{hourglass}, and RGB-D PoseNet \cite{dualstream_posenet}. We also compare the results to the depth-only PoseNet and PointNet-Pose data that we collected. 


\subsection{Results}
Based on \hyperref[tab:results]{Table II}, FusionLoc performs better than a number of other camera relocalization methods tested. The visualized results can be seen in \hyperref[fig:graphs]{Figs. 3-5}, where the scenes are in the same order as the table. Perhaps the most important method to compare against is RGB-D PoseNet \cite{dualstream_posenet}, which also has a dual stream architecture. Notice PoinNet-Pose alone has a better average performance compared to RGB-D PoseNet. While some of the improvement can be attributed to a different loss function, the change in depth representation plays a vital role, reducing the overall pose error. In fact, interestingly, PointNet-Pose is comparable to many of the other methods as well, without using any RGB data.\\
\indent FusionLoc on the other hand, achieves accuracies that either perform better, or are comparable to the baseline methods. A dataset where we expected better performance however, is the Stairs dataset. Pseudo-LiDAR is able to to capture the change in depth of the staircase more accurately which should have translated to more accurate pose estimation. Upon further inspection of the dataset, we believe this may be due to the inconsistencies found in the depth maps, specifically the missing depth information behind the stairs. While such patches exist in other scenes as well, they seem to be consistent throughout the Stairs scene. Regardless, FusionLoc still achieves comparable performance on the Stairs scene as well.\\
\indent Overall, pseudo-LiDAR proves to be a better alternative to depth maps for localization. We also believe that FusionLoc can achieve even better accuracies with psudo-LiDAR signals generated from improved depth maps. \\

\section{Conclusion}

In this paper, we present an argument for the viability of pseudo-LiDAR for localization purposes in order to create more accurate large scale systems for applications such as autonomous vehicles, mobile robots, and augmented reality. Further, we present results from FusionLoc, a novel architecture that uses pseudo-LiDAR and RGB images to regress a 6DOF camera pose, as well as PointNet-Pose, a PointNet inspired localization network that processes exclusively on pseudo-LiDAR data. \\
\indent With the many camera localization techniques proposed in the past, we select various competing and state-of-the-art methods to use as a baseline in our experiments and compare against FusionLoc. In general, using pseudo-LiDAR over depth maps for localization proves to be more accurate due to object shapes and sizes being preserved when the point clouds are processed.\\
\indent While FusionLoc is competitive with other camera pose regression methods, we believe there is still room for improvement and further testing. We can incorporate muti-scale grouping \cite{qi2017pointnetplusplus} in the SA layers for more robust features. Furthermore, it may also be beneficial to add temporal constraints to learn features consistent throughout pointsets and images, similar to MapNet \cite{brahmbhatt2018geometryaware}. Finally, we believe FusionLoc can produce even better results with improved depth maps. 7 Scenes was collected using the Kinect v1, however, since then, depth cameras and depth estimation techniques, including monocular depth estimation, have seen significant improvements. We would like to further test FusionLoc using depth maps generated using varying techniques and tools.\\

\addtolength{\textheight}{-12cm}   


\section*{Acknowledgment}
This research was supported by the National Sciences and Research Council of Canada (NSERC) grant USRA/567510-2021, the University of Toronto Mississauga's (UTM) Office of the Vice-Principal Research (OVPR) Fund, Dean's Office Research Opportunity Project (ROP) Fund, and UTM's Department of Mathematical and Computational Sciences.\\


\balance
\bibliographystyle{IEEEtran}
\bibliography{references}

\end{document}